\documentclass[preprint,3p]{elsarticle}
\usepackage{epstopdf}
\usepackage{caption}
\usepackage{setspace}
\usepackage{amssymb}
\usepackage{multirow}
\usepackage{amsthm}
\usepackage{booktabs}
\usepackage{amsmath}
\usepackage{amssymb}
\usepackage{mathrsfs}
\usepackage{longtable}
\usepackage{lscape}
\usepackage{url}
\usepackage{tabularx}
\usepackage{epsfig}
\usepackage{colortbl}
\usepackage{subfigure}
\usepackage{tikz,mathpazo}
\usepackage{graphicx}
\usepackage{natbib}
\usepackage[noend]{algpseudocode}
\usepackage{algorithm}
\usepackage{algorithmicx}
\newtheorem{myDef}{Definition}

\usetikzlibrary{shapes.geometric, arrows}
\biboptions{numbers,sort&compress}

\begin{document}
\begin{frontmatter}
\title{TDQMF: Two-dimensional quantum mass function}
\author[address1]{Yuanpeng He}
\author[address1]{Fuyuan Xiao \corref{label1}}

\address[address1]{School of Computer and Information Science, Southwest University, Chongqing, 400715, China}
\cortext[label1]{Corresponding author: Fuyuan Xiao, School of
Computer and Information Science, Southwest University, Chongqing,
400715, China. Email address: xiaofuyuan@swu.edu.cn,
doctorxiaofy@hotmail.com.}
\begin{abstract}
Quantum mass function has been applied in lots of fields because of its efficiency and  validity of managing  uncertainties in the form of quantum which can be regarded as an extension of classical Dempster‐Shafer (D‐S) evidence theory. However, how to handle uncertainties in the form of quantum is still an open issue. In this paper, a new method is proposed to dispose uncertain quantum information, which is called two-dimensional quantum mass function (TDQMF). A TDQMF is consist of two elements, $TQ = (\mathbb{Q}_{original},\mathbb{Q}_{indicative})$, both of the $\mathbb{Q}s$ are quantum mass functions, in which the $\mathbb{Q}_{indicative}$ is an indicator of the reliability on $\mathbb{Q}_{original}$. More flexibility and effectiveness are offered in handling uncertainty in the field of quantum by the proposed method compared with primary quantum mass function. Besides, some numerical examples are provided and some practical applications are given to verify its correctness and validity.
\end{abstract}
\end{frontmatter}
\section{Introduction}
It is unavoidable to deal with indeterminate information in practical situations \cite{Deng2020ScienceChina, Seiti2018, MURPHY20001, gao2019uncertaintyIJIS, Wang2018uncertainty}. Lots of researches have been made to seek for truly useful information contained in uncertainty \cite{meng2018fluid, deng2019zero, Yager2018}. Therefore, to address this problem, many methods have been proposed, such as $Z$-numbers \cite{Zadeh2011, li2020newuncertainty, Jiang2019Znetwork, liu2019derive}, $D$-numbers \cite{Deng2012, Deng2014, Xiao2019a, deng2019evaluating, IJISTUDNumbers}, probability theory \cite{Yager2014, Zhang2020b, Jiang2019IJIS}, soft sets \cite{Feng2018, Feng2016}, Dempster‐Shafer evidence theory \cite{Dempster1967Upper, book}, complex evidence theory \cite{Xiao2019complexmassfunction, Xiao2020b, Xiao2020CEQD, Xiao2020maximum},evidential reasoning \cite{Jiang2019, Zhou2018, Liu2018a, liao2020deng}, fuzzy sets \cite{Zadeh1965,Zadeh1979, Xue2020entailmentIJIS, 8944285, Zhou2020} and quantum mass function \cite{Gao2019quantummodel, Dai2020, Huang2019}.

Among those proposed methods, quantum mass function \cite{Gao2019quantummodel} is the most widely useful and novel tool to solve problems occurred in traditional evidence theory at the level of quantum. In general, it is common to meet with a variety of applications developed on the base of classic evidence theory, like pattern recognition \cite{Liu2019, Song2018}. However, with respect to solving conflicts and dispelling uncertainties at the level of quantum, there is no any other previous methods to figure out the answer to related issues. In other words, researchers are not able to find reliable and effective improved methods to handle different kinds of troublesome errors appeared in the process of utilizing quantum mass function to dispose information in the form of quantum. 

Therefore, in this paper, a specific and customized method is proposed to avoid appearances of various problems when solving relevant questions at the level of quantum. Two dimensions of quantum basic probability assignment (QBPA) are designed to improve reliability and availability of original quantum mass function, which is called Two-dimensional quantum mass function (TDQMF) and denoted as $\mathbb{T} = \{\mathbb{Q}_{1},\mathbb{Q}_{2}\}$. The first dimension of TDQMF is consist of original QBPAs whose concrete constitution, namely frame of discernment in the form of quantum changes according to actual situations. Besides, the second one is a measure of confidential level of the first dimension whose essences are still QBPAs, but the difference is that the first dimension focuses on practical affairs in real life while the second one puts attention on QPBAs contained in the first dimension themselves which dose not considers the whole system of judgements in the way of the first dimension. And the frame of discernment of the second dimension is certain which is defined as $\Theta = \{\emptyset, Y, N, YN\}$. Among them, proposition "$Y$" represents "approval", "$N$" means "disagreement" and "$YN$" denotes that no further information can be provided. Besides, the contributions of TDQMF can be listed as: 
\begin{itemize}
\item[$\bullet$] TDQMF is a pioneer in amending the disposing process involved in basic probability assignment in the form of quantum which provides a convenient way improve the reliability and validity of judgements given by original quantum mass function.
\end{itemize}
\begin{itemize}
	\item[$\bullet$] A completely novel rule of combination of TDQMF is proposed to provide credible results about real situations.
\end{itemize}
\begin{itemize}
	\item[$\bullet$] A corresponding algorithm developed on the base of TDQMF is designed to handle separate kinds of conditions and specific problems brought by practical applications.
\end{itemize}
All in all, TDQMF has the ability to handle problems in different fields of applications in the form of quantum and creates a new direction to manage quantum information.

The rest of this paper is organized as follows. The section of preliminaries introduces concepts of quantum mass function \cite{Gao2019quantummodel} and $Z$-numbers \cite{Zadeh2011}. In the section of proposed method, the definition of TDQMF is given and a relevant rule of combination of it is also devised. Besides, for the next section, 4 applications are provided to verify the correctness and effectiveness of TDQMF involved in 4 diverse aspects: target recognition, decision making, income estimate and fault diagnosis. Conclusions are given in the last section.

\section{Preliminaries}
\label{Preliminaries}
In this section, some related concepts including quantum mass function and $Z$-numbers are briefly introduced. 
\subsection{Quantum mass function}
\begin{myDef}(Quantum frame of discernment)\end{myDef}
Assume $\Theta$ is an exclusive and non-empty set whose elements $|A_{i} \rangle $ are mutually exclusive and consistent. Therefore, set $\Theta$ is named as the quantum frame of discernment. $\Theta$ is defined as \cite{Gao2019quantummodel}:
\begin{equation}
|\Theta \rangle = \{|A_{1} \rangle, |A_{2} \rangle, |A_{3} \rangle, ... , |A_{i} \rangle, ... , |A_{n} \rangle\}
\end{equation}
\par And the power set of $|\Theta \rangle$ which is consist of $2^{N}$ separate elements is defined as \cite{Gao2019quantummodel}:
\begin{equation}
2^{|\Theta \rangle} = \{\emptyset, |A_{1} \rangle, |A_{2} \rangle, ..., |A_{n} \rangle, ..., \{|A_{1} \rangle, |A_{2} \rangle\}, ... , |\Theta \rangle\}
\end{equation}
\begin{myDef}(Quantum mass function)\end{myDef}
On the base of the definition of the quantum frame of discernment, the quantum mass function $\mathbb{Q}$ can be defined as \cite{Gao2019quantummodel}:
\begin{equation}
\mathbb{Q}(|A \rangle) = \psi e^{j\theta}
\end{equation}
\par And it satisfies the properties which are defined as \cite{Gao2019quantummodel}:
\begin{equation}
\mathbb{Q}:2^{|\Theta \rangle} \rightarrow [0,1]
\end{equation}
\begin{equation}
\mathbb{Q}(\emptyset) = 0
\end{equation}
\begin{equation}
\sum_{\mathbb{Q}(|A \rangle) \subseteq \mathbb{Q}(|\Theta \rangle)}|\mathbb{Q}(|A \rangle)| = 1
\end{equation}
\par where the value of $|\mathbb{Q}(|A \rangle)|$ is equal to $\psi^{2}$. And the quantum mass function is also called quantum basic probability assignment (QBPA), in which the mass of $\psi^{2}$ represents the belief level of support of certain quantum proposition $|A \rangle$.
\begin{myDef}(Quantum rule of combination)\end{myDef}
\par Assume there are 2 QBPAs $\mathbb{Q}_{1}$ and $\mathbb{Q}_{2}$ and the rule of combination is defined as follows \cite{Gao2019quantummodel}:
\begin{equation}
\mathbb{Q}(|A \rangle) = \left\{
\begin{aligned}
\frac{1}{1 - |\mathbb{K}|} \sum_{|B \rangle \cap |C \rangle = |A \rangle} \mathbb{Q}(|B \rangle) \times \mathbb{Q}(|C \rangle) \ \ \ \  |A \rangle \neq \emptyset \\ 
0  \ \ \ \ \ \ \ \ \ \ \ \ \ \ \ \ \ \ \ \ \ \ \ \ \ \ \ \ \ \ \ \  |A \rangle = \emptyset
\end{aligned}
\right.
\end{equation}
\par The step of normalization is defined as \cite{Gao2019quantummodel}:
\begin{equation}
|\mathbb{Q}(|A \rangle)| = \frac{|\mathbb{Q}(|A \rangle)|}{|\mathbb{Q}(|A \rangle)| + |\mathbb{Q}(|B \rangle)|+ ... + |\mathbb{Q}(|A \rangle,|B \rangle)|+ ...}
\end{equation}
\par Besides, the factor of normalization is defined as \cite{Gao2019quantummodel}:
\begin{equation}
\mathbb{K} = \sum_{|B \rangle \cap |C \rangle = \emptyset} \mathbb{Q}_{1}(|B \rangle) \times \mathbb{Q}_{2}(|C \rangle)
\end{equation}
\par In the rule of combination of quantum, the factor $\mathbb{K}$ can be regarded as a kind of probability of quantum and $|\mathbb{K}|$ is a degree which measures the conflict among QBPAs.
\subsection{Z-numbers}
Z-numbers is a kind of pioneering method to measure uncertainties existing in information, which takes reliability of information into consideration. As some higher requirements of disposing information have been promoted, reliability of information has been paid more and more attention and focus \cite{Zadeh2011}.
\begin{myDef}A Z-number is consist of two fuzzy numbers and its basic form is defined as Z = (A,B). Assume there is a kind of uncertain variable, X, a corresponding Z-number is connected with it. The first part of the constitution of the Z-number, A, is denoted as an estimate of variable X. More than that, the second part of the constitution of the Z-number, B, is a measure of reliability of the first component, A.\end{myDef}
\section{Proposed method: TDQMF: two-dimensional quantum mass function}
In this section, some relevant statement and definitions are provided. On the basis of information given, the rule of combination with respect to TDQMF is developed.
\subsection{The definition of TDQMF}
Assume $\mathbb{Q}_{1}$ and $\mathbb{Q}_{2}$ are two QBPAs, the second one is a measure of the degree of reliability of the first one. Thus, A TDQMF is defined as:
\begin{equation}
\mathbb{T} = \{\mathbb{Q}_{1},\mathbb{Q}_{2}\}
\end{equation}
And the properties it satisfies are defined as: \par 1. The frame of discernment in the form of quantum for $\mathbb{Q}_{1}$ is $2^{|\Theta \rangle} = \{\emptyset, |\mathbb{Q}_{1} \rangle, |\mathbb{Q}_{2} \rangle, ..., |\mathbb{Q}_{n} \rangle, ..., \{|\mathbb{Q}_{1} \rangle, |\mathbb{Q}_{2} \rangle\}, \\... , |\Theta \rangle\}$. According to the definition of the frame of discernment, the mass of $|\mathbb{Q}_{1} \rangle$ represents the most immediate support of the quantum proposition $|\mathbb{Q}_{1} \rangle$.\par 2. The frame of discernment in the form of quantum for $\mathbb{Q}_{2}$ is defined as $\Theta = \{\emptyset, Y, N, YN\}$, in which $"Y"$ denotes $"support"$ to $\mathbb{Q}_{1}$ and $"N"$ denotes $"not\ support"$ to $\mathbb{Q}_{1}$. In other words, the value of proposition $"Y"$ represents the degree of the reliability of $\mathbb{Q}_{1}$ and the value of proposition $"N"$ represents the degree of the unreliability of $\mathbb{Q}_{1}$. Besides, proposition $"YN"$ manifests uncertainty and hesitation in measuring whether $\mathbb{Q}_{1}$ is reliable or unreliable. More than that, if the mass of $\mathbb{Q}_{1}$ is very close to actual conditions, the proposition $"Y"$ in the frame of discernment of $\mathbb{Q}_{2}$ can obtain a much higher value and the proposition $"N"$ is expected to get a much lower value. In the opposite, if the mass of $\mathbb{Q}_{1}$ is distant from actual conditions, the proposition $"Y"$ in the frame of discernment of $\mathbb{Q}_{2}$ can obtain a much lower value and the proposition $"N"$ is expected to get a much higher value. It can be easily concluded that, the second dimensional judgement provides a much more convenient way to guarantee enough accuracy in decision making and pattern recognition.\\

\setlength{\tabcolsep}{4mm}{\begin{table}[h]
		\centering
		\caption{The frame of discernment for $\mathbb{Q}_{1}$ in Example 1.}
		\begin{spacing}{1.40}
			\begin{tabular}{l c c c c c}\hline
				$\mathbb{Q}_{1}$ &$\{x_{1}\}$& $\{x_{2}\}$&$\{x_{2}, x_{3}\}$\\\hline 
				&$0.7071e^{0.7854j}$&$0.2828e^{0.7854j}$&$0.6782e^{0.4850j}$\\\hline  
			\end{tabular}
		\end{spacing}
\end{table}}
\setlength{\tabcolsep}{4mm}{\begin{table}[h]
		\centering
		\caption{The frame of discernment for $\mathbb{Q}_{2}$ in Example 1.}
		\begin{spacing}{1.40}
			\begin{tabular}{l c c c c c}\hline
				$\mathbb{Q}_{2}$ &$\{Y\}$& $\{N\}$&$\{YN\}$\\\hline 
				&$0.8306e^{0.7854j}$&$0.5385e^{0.7854j}$&$0.1414e^{0.7854j}$\\\hline  
			\end{tabular}
		\end{spacing}
\end{table}}
\textbf{Example 1:} Assume the frame of discernment in the form of quantum for $\mathbb{Q}_{1}$ is $\Theta = \{x_{1},x_{2},x_{3}\}$, the propositions existing in the frame of discernment and their corresponding values are provided in Table 1. More than that, assume there are 100 experts and 69 of them approve the judgements made by $\mathbb{Q}_{1}$. Besides, the remain of them hold the opposite opinion except 2 experts can not made final decisions. Then, it can be easily obtained the values of the propositions existing in the frame of discernment of $\mathbb{Q}_{2}$. And the values are given in Table 2.
\par \textbf{Note: }In order to simplify the process of transferring real numbers into the form of quantum (if needed), the value is supposed to be divided equally and then allocated to real part and imaginary part under certain regulations.
\subsection{The rule of combination of TDQMF}
In the last part, it can be easily summarized that the second dimension of judgements could play a very important role in properly adjusting the values of the propositions existing in certain frames of discernment from different evidences to obtain reasonable and valid results about estimates on practical circumstances. Therefore, the needs of designing a special method to combine TDQMF are urgent. In this section, an innovative rule of combination is proposed to produce accurate and valid judgements on practical situations.
\begin{equation}
\mathbb{Q}_{final}(\{x_{i}\}) = (\mathbb{Q}_{1}(\{x_{i}\})   + \mathbb{Q}_{1}(\{X_{x_{i}}\}) \times \frac{1}{n}) \times \mathbb{Q}_{2}(\{Y\}) + \mathbb{Q}_{1}(\{X_{non-x_{i}}\}) \times \mathbb{Q}_{2}(\{N\}) \ \ \ \ x_{i} \subset 2^{\Theta}
\end{equation}
\begin{equation}
\mathbb{Q}_{final}(\{X_{x_{i}}\}) = \frac{n^{2} - n}{n^{2}} \times \mathbb{Q}_{1}(\{X_{x_{i}}\}) \times \mathbb{Q}_{2}(\{Y\})\ \ \ \ x_{i} \subset 2^{\Theta}\ \ and\ \ x_{i} \neq \Theta
\end{equation}
\begin{equation}
\mathbb{Q}_{final}(\{{\Theta}\}) = \frac{n^{2} - n}{n^{2}} \times \mathbb{Q}_{1}(\{\Theta\}) + \mathbb{Q}_{2}(\{YN\})
\end{equation}
\begin{figure}
	\centering
	\includegraphics[width=0.8\linewidth]{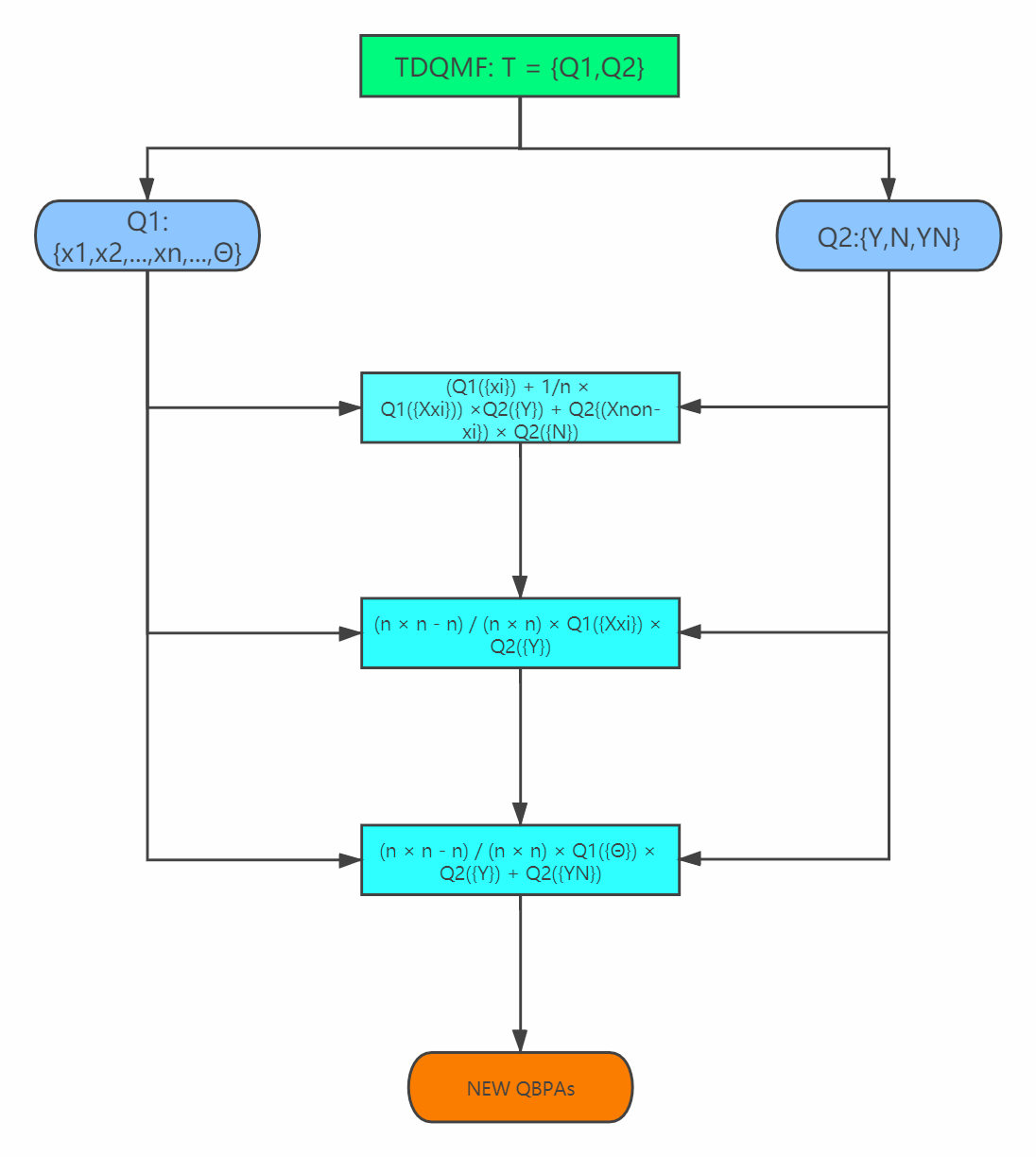}
	\caption{The procedure of combination of TDQMFs. Details of each step are offered in this flow chart.}
	\label{fig:1}
\end{figure}
\par Where $\{X_{x_{i}}\}$ and $\{X_{non-x_{i}}\}$ represent multiple propositions. And subscripts $x_{i}$ and $non-x_{i}$ means proposition $x_{i}$ and $non-x_{i}$ are part of the monolithic proposition $X$. In the construction of the rule of combination, for maximized utilization of given information, multiple propositions are divided proportionally according to the number of single propositions which are contained in this multiple proposition and distributed to single propositions which takes advantage of an efficient method called pignistic transformation \cite{article} which reduces uncertainty in handling ambiguous evidences. Besides, a negation of a negation of certain propositions is regarded as an approval of certain propositions. Therefore, with respect to single propositions, their mass is consist of the values of a direct approval to certain propositions and a dual negation of them. However, it is not proper to eliminate the values of multiple propositions because all of the values are managed in the form of quantum. In order to reduce uncertainty existing in the frame of discernment but not lose accuracy in describing actual conditions, it is proposed that $\frac{n^{2} - n}{n^{2}}$ of the values of multiple propositions are kept due to the definition of the quantum mass function, namely a value of quantum is a sum of squares of the numbers of real part and imaginary part. More than that, a dual negation of multiple propositions is not taken into consideration because reducing uncertainties is the main goal in managing quantum information. In the last, universal sets of frames of discernment are disposed solely because the second dimensional judgement may contain different levels of hesitation and all of information provided is supposed to be utilized to illustrate practical situations in a more sufficient way.
\par After those modifications, all of the values obtained are expected to be combined using quantum rule of combination $n - 1$ times to get final descriptions of circumstances (n is the number of evidences). 
\par However, after a reallocation of the mass, a step of normalization to guarantee that the sum of all the values of propositions existing in frames of discernment is exactly equal to 1 is needed. A part of the  operation of normalization is defined as:
\begin{equation}
|\mathbb{Q}_{FINAL}(\{{X_{i}}\})| = \frac{|\mathbb{Q}_{final}(\{{X_{i}}\})|}{|\mathbb{Q}_{final}(\{{X_{1}}\})| + |\mathbb{Q}_{final}(\{{X_{2}}\})| + ... + |\mathbb{Q}_{final}(\{{X_{i}}\})| + ... + |\mathbb{Q}_{final}(\{{X_{k}}\})|}   \ \ \ \ k = 2^{\Theta}
\end{equation}
According to the value obtained, $|\mathbb{Q}_{FINAL}(\{{X_{i}}\})|$, it can be easily calculated to get the ratio of original values and normalized ones, $\xi$. Therefore, real part and imaginary part of QBPAs can be adjusted on the basis of the extraction of a root, $\xi$, in the light of definition of QBPAs.
\par \textbf{Note: }When the value of $\mathbb{Q}_{2}(\{Y\})$ is equal to 1, modified values of singletons is equal to the ones modified by pignistic transformation \cite{article} at the level of quantum.
\subsection{Algorithm developed on the base of TDQMF}
\par \textbf{Step1: }Assume there is a group of TDQMFs, $\mathbb{T} = \{\mathbb{Q}_{1},\mathbb{Q}_{2}\}$. According to the proposed rule of combination of TDQMF, a combined quantum evidence, $\mathbb{Q}_{final}$, can be easily obtained.
\par \textbf{Step2: }In order to ensure the sum of value of every proposition existing in the frame of discernment is equal to 1, a step of normalization is expected to be carried out. And what should be pointed out is that values which are calculated are modulo mass of each proposition not the ones which are in the form of quantum.
\par \textbf{Step3: }Using the rule combination of quantum mass function, a final quantum evidence can be obtained. The process of combination can be briefly defined as:
\begin{equation}
\mathbb{Q}_{FINAL}(\{X_{i}\}) = (((Q_{1}(\{X_{i}\}) \otimes Q_{2}(\{X_{i}\})) \otimes Q_{3}(\{X_{i}\})) \otimes ...)
\end{equation}
\par \textbf{Step4: }According to definition of quantum mass function, calculate modulo value of $\mathbb{Q}_{FINAL}(\{X_{i}\}) \ \ (0 \leq i \leq n)$. 
\par \textbf{Step5: }Find the proposition whose modulo value is the biggest among all the values of other propositions.
\begin{equation}
|\mathbb{Q}_{SELECTED}| = MAX{|\mathbb{Q}_{FINAL}(\{x_{i}\})|} \ \ (0 \leq i \leq n)
\end{equation}
\par \textbf{Step6: }There are two kinds of situations and two different standard of judgement can be set therefore. In the first case, the proposition which has the biggest value is directly selected as the target without having any other operations. In the second case, a threshold $\curlyvee$ can be established to serve as a kind of standard in choosing proper target. If $|\mathbb{Q}_{SELECTED}| \geq \curlyvee$, then the proposition whose modulo value is equal to $|\mathbb{Q}_{SELECTED}|$ is selected as target.
\begin{equation}
\mathbb{Q}_{SELECTED} = \{x_{i}\}
\end{equation}
\begin{equation}
Target \leftarrow \{x_{i}\}
\end{equation}
 \begin{algorithm}
	\caption{:The details of the proposed algorithm } 
	\textbf{Input:} A group of TDQMFs which are defined as $\mathbb{T}_{j} = \{\mathbb{Q}_{1},\mathbb{Q}_{2}\}\ \ (0 \leq j \leq n)$\\ 
	A frame of discernment in the form of quantum$|\Theta \rangle = \{|A_{1} \rangle, |A_{2} \rangle, |A_{3} \rangle, ... , |A_{i} \rangle, ... , |A_{n} \rangle\}$\\
	A frame of discernment of the second dimension of judgement $\Theta = \{\emptyset, Y, N, \Theta\}$\\
	\textbf{Output:} The information of the chosen target\\
	$\textbf{for} \ j =1; \ j \leq n$ \ \textbf{do}  \\
	1. Combine different two dimensional quantum evidences $(\mathbb{T}_{j})$ to get the modified QBPAs using rule of combination of TDQMF.\\
	\textbf{end}\\
	2. Normalize every component contained in the modified QBPAs.\\
	3. Combine QBPAs using rule of combination of QBPAs to get final results.\\
	4. Calculate modulo values of different propositions contained in the final results.\\
	5. Select the proposition whose value is the biggest.
	--- $MAX = ||A_{1} \rangle|$\\
	------ $\textbf{for} \ i =1; \ i \leq n$ \ \textbf{do}  \\
	---------  $\textbf{if}\ \ ||A_{i} \rangle|  > MAX$\\
	------------ $MAX = ||A_{i} \rangle|$\\
	------ \textbf{end}\\
	6.\textcircled{{\footnotesize 1}}: If there is no other limitation, the proposition which corresponds with the value of $MAX$ is selected as target.
	
	\ \ \  \textcircled{{\footnotesize 2}}: If a threshold is set, only when the value of $||A_{i} \rangle|$ is bigger than threshold, a proposition can be chosen as target if its mass is equal to the value of $||A_{i} \rangle|$. In the opposite, no decisions can be made.
\end{algorithm}

 On the contrary, if $|\mathbb{Q}_{SELECTED}| \leq \curlyvee$, then no target can be chosen, which means no substantial decisions can be made.
\clearpage
\section{Applications}
\subsection{Application of target recognition}
In real world, there are lots of uncertainties which leads to chaos when coming to trying to figure out what are the practical situations. In order to solve problems brought by uncertainties, researchers have made a lot efforts in identifying proper targets \cite{LiuF2020TFS, Han2018, Pan2020association, han2018evidential, deng2010target}. However, how to recognize corresponding objects rightly at the level of quantum is still an open issue. The following application illustrates the efficiency and validity of TDQMF in target recognition.
\subsubsection{Application 1}
Assume there is a system which has different sources of information from multiple transducers which is utilized for target recognition. And three targets are the objects which this system tries to identify. Therefore, the frame of discernment of the first dimension can be defined as $\Theta = \{x_{1}, x_{2}, x_{3}, x_{1}x_{2}\}$ and frame of discernment of the second dimension is defined as $\Theta = \{Y, N, YN\}$. Besides, a hypothesis is that there is only one target can be selected and four pieces of TDQMF, $\mathbb{T} = \{\mathbb{Q}_{1},\mathbb{Q}_{2}\}$ are provided. The details of first dimension of TDQMF are given in Table 3. And the details of second dimension of TDQMF are given in Table 4. More than those, exhaustive details of calculation process are provided in this first application.
\begin{table}[h]\footnotesize
	\centering
	\caption{Quantum evidences given by multiple sensors}
	\begin{spacing}{1.40}
		\begin{tabular}{c c c c c}\hline
			$Quantum \ \ evidences$ &\multicolumn{4}{c}{$Values \ \ of \ \ propositions$}\\\hline
			&$\{x_{1}\}$&$\{x_{2}\}$&$\{x_{3}\}$&$\{x_{1}, x_{2}\}$\\
			$\mathbb{Q}_{1}^{1}$&$0.5000e^{0.9271j}$&$0.6403e^{0.8960j}$&$0.4472e^{0.4636j}$&$0.3741e^{1.0068j}$\\  
			$\mathbb{Q}_{1}^{2}$&$0.5385e^{1.1902j}$&$0.7071e^{0.7853j}$&$0.3162e^{0.3217j}$&$0.3316e^{0.4404j}$\\
			$\mathbb{Q}_{1}^{3}$&$0.5047e^{0.9827j}$&$0.6224e^{0.8081j}$&$0.4640e^{0.6477j}$&$0.3774e^{1.3026j}$\\
			$\mathbb{Q}_{1}^{4}$&$0.4609e^{0.8621j}$&$0.6440e^{0.9397j}$&$0.4070e^{0.4855j}$&$0.4549e^{0.4964j}$\\\hline
		\end{tabular}
	\end{spacing}
\end{table}
\begin{table}[h]\footnotesize
	\centering
	\caption{Experts' judgements on reliability of quantum evidences given by sensors}
	\begin{spacing}{1.40}
		\begin{tabular}{c c c c c c}\hline
			$Quantum \ \ judgements$ &\multicolumn{4}{c}{$Values \ \ of \ \ propositions$}\\\hline
			&$\{Y\}$& $\{N\}$&$\{Y, N\}$&$\{\emptyset\}$\\
			$\mathbb{Q}_{2}^{1}$&$0.8366e^{0.2970j}$&$0.4472e^{0.4636j}$&$0.3162e^{0.3217j}$&$0$\\  
			$\mathbb{Q}_{2}^{2}$&$0.7745e^{0.4423j}$&$0.4690e^{0.5493j}$&$0.4242e^{0.6004j}$&$0$\\
			$\mathbb{Q}_{2}^{3}$&$0.8062e^{0.1243j}$&$0.5099e^{0.7587j}$&$0.3000e^{0.8410j}$&$0$\\
			$\mathbb{Q}_{2}^{4}$&$0.7681e^{0.4242j}$&$0.5477e^{0.7519j}$&$0.3316e^{1.0764j}$&$0$\\\hline   
		\end{tabular}
	\end{spacing}
\end{table}

First, it is supposed to use the rule of combination of TDQMF to obtain modified QBPAs. And selecting corresponding propositions correctly is the most important.

For example, the process of obtaining the modified results of the first piece of quantum evidence is given as:

$\mathbb{Q}_{FINAL_{1}}^{1}(\{x_{1}\}) = (\mathbb{Q}_{1}^{1}(\{x_{1}\}) + \mathbb{Q}_{1}^{1}(\{x_{1}, x_{2}\}) \times \frac{1}{n}) \times \mathbb{Q}_{2}^{1}(\{Y\}) +  \mathbb{Q}_{1}^{1}(\{X_{non-x_{1}}\}) \times \mathbb{Q}_{2}^{1}(\{N\}) = (0.5000e^{0.9271j} + 0.3741e^{1.0068j} \times \frac{1}{2}) \times 0.8366e^{0.2970j} + (0.6403e^{0.8960j} + 0.4472e^{0.4636j}) \times 0.4472e^{0.4636j} = 1.0493e^{1.2172j}$

$\mathbb{Q}_{FINAL_{1}}^{1}(\{x_{2}\}) = (\mathbb{Q}_{1}^{1}(\{x_{2}\}) + \mathbb{Q}_{1}^{1}(\{x_{1}, x_{2}\}) \times \frac{1}{n}) \times \mathbb{Q}_{2}^{1}(\{Y\}) +  \mathbb{Q}_{1}^{1}(\{X_{non-x_{2}}\}) \times \mathbb{Q}_{2}^{1}(\{N\}) = (0.6403e^{0.8960j} + 0.3741e^{1.0068j} \times \frac{1}{2}) \times 
0.8366e^{0.2970j} + (0.5000e^{0.9271j} + 0.4472e^{0.4636j}) \times 0.4472e^{0.4636j} = 1.0708e^{1.2290j}$

$\mathbb{Q}_{FINAL_{1}}^{1}(\{x_{3}\}) = (\mathbb{Q}_{1}^{1}(\{x_{3}\})) \times \mathbb{Q}_{2}^{1}(\{Y\}) +  \mathbb{Q}_{1}^{1}(\{X_{non-x_{3}}\}) \times \mathbb{Q}_{2}^{1}(\{N\}) =  (0.4472e^{0.4636j}) \times 0.8366e^{0.2970j} + (0.5000e^{0.9271j} + 0.6403e^{0.8960j} + 0.3741e^{1.0068j}) \times 0.4472e^{0.4636j} = 1.0024e^{1.1736j}$  

$\mathbb{Q}_{FINAL_{1}}^{1}(\{x_{1}, x_{2}\}) = \frac{n^{2} - n}{n^{2}} \times \mathbb{Q}_{1}^{1}(\{x_{1}, x_{2}\}) \times \mathbb{Q}_{2}^{1}(\{Y\}) = \frac{2^{2} - 2}{2^{2}} \times 0.3741e^{1.0068j} \times 0.8366e^{0.2970j} = 0.1565e^{1.3038j}$

$\mathbb{Q}_{FINAL_{1}}^{1}(\{\Theta\}) = \frac{n^{2} - n}{n^{2}} \times \mathbb{Q}_{1}^{1}(\{\Theta\}) + \mathbb{Q}_{2}^{1}(\{\Theta\}) = 0 + 0.3162e^{0.3217j} = 0.3162e^{0.3217j}$

Besides, the modified results of all of the propositions are given in Table 5.
\begin{table}[h]\footnotesize
	\centering
	\caption{Quantum evidences after modification}
	\begin{spacing}{1.40}
		\begin{tabular}{c c c c c c}\hline
			$Combined \ \ TDQMF$ &\multicolumn{5}{c}{$Values \ \ of \ \ propositions$}\\\hline
			&$\{x_{1}\}$&$\{x_{2}\}$&$\{x_{3}\}$&$\{x_{1}, x_{2}\}$&$\{\Theta\}$\\
			$\mathbb{Q}_{Modified_{1}}^{1}$&$1.0493e^{1.2172j}$&$1.0708e^{1.2290j}$&$1.0024e^{1.1736j}$&$0.1565e^{1.3038j}$&$0.3162e^{0.3217j}$\\
			$\mathbb{Q}_{Modified_{1}}^{2}$&$0.9785e^{1.3347j}$&$1.0281e^{1.2555j}$&$0.9202e^{1.2418j}$&$0.1283e^{0.8830j}$&$0.4242e^{0.6004j}$\\
			$\mathbb{Q}_{Modified_{1}}^{3}$&$1.0928e^{1.3458j}$&$1.0874e^{1.2762j}$&$1.0108e^{1.4370j}$&$0.1522e^{1.4270j}$&$0.3000e^{0.8410j}$\\
			$\mathbb{Q}_{Modified_{1}}^{4}$&$1.0661e^{1.3481j}$&$1.1190e^{1.3277j}$&$1.1071e^{1.3731j}$&$0.1072e^{1.2313j}$&$0.3316e^{1.0764j}$\\\hline  
		\end{tabular}
	\end{spacing}
\end{table}

Second, after obtaining modified QBPAs, a step of normalization is supposed to be executed to ensure that the sum of those modified QBPAs is still equal to 1 and meets the requirement of the definition of quantum mass function. Besides, all of the normalized values of QBPAs are listed in Table 6.
\begin{table}[h]\footnotesize
	\centering
	\caption{Quantum evidences after normalization}
	\begin{spacing}{1.40}
		\begin{tabular}{c c c c c c}\hline
			$Normalized \ \ quantum \ \ evidences$ &\multicolumn{5}{c}{$Values \ \ of \ \ propositions$}\\\hline
			&$\{x_{1}\}$&$\{x_{2}\}$&$\{x_{3}\}$&$\{x_{1}, x_{2}\}$&$\{\Theta\}$\\
			$\mathbb{Q}_{Normalized_{1}}^{1}$&$0.5650e^{1.2172j}$&$0.5942e^{1.2010j}$&$0.5398e^{1.1736j}$&$0.0842e^{1.3038j}$&$0.1702e^{0.3217j}$\\  
			$\mathbb{Q}_{Normalized_{1}}^{2}$&$0.5595e^{1.3347j}$&$0.5879e^{1.2555j}$&$0.5262e^{1.2418j}$&$0.0734e^{0.8830j}$&$0.2425e^{0.6004j}$\\   
			$\mathbb{Q}_{Normalized_{1}}^{3}$&$0.5831e^{1.3459j}$&$0.5803e^{1.2762j}$&$0.5393e^{1.4370j}$&$0.0812e^{1.4270j}$&$0.1600e^{0.8410j}$\\ 
			$\mathbb{Q}_{Normalized_{1}}^{4}$&$0.5523e^{1.3481j}$&$0.5797e^{1.3731j}$&$0.5735e^{1.3731j}$&$0.0540e^{1.3218j}$&$0.1637e^{1.0764j}$\\\hline  
		\end{tabular}
	\end{spacing}
\end{table}

Third, combination of QBPAs is expected to be carried out according to the rule of combination of quantum mass function to obtain the final values of indicator of targets. All of the combined results are given in Table 7.
\begin{table}[h]\footnotesize
	\centering
	\caption{Quantum evidence after combination}
	\begin{spacing}{1.40}
		\begin{tabular}{c c c c c c}\hline
			$Combined \ \ quantum \ \ evidence$ &\multicolumn{5}{c}{$Values \ \ of \ \ propositions$}\\\hline
			&$\{x_{1}\}$&$\{x_{2}\}$&$\{x_{3}\}$&$\{x_{1}, x_{2}\}$&$\{\Theta\}$\\
			$\mathbb{Q}_{Combined_{1}}$&$0.6140e^{1.3082j}$&$0.6918e^{1.1929j}$&$0.3799e^{1.2478j}$&$0.004e^{0.2703j}$&$0.0016e^{-0.5758j}$\\\hline  
		\end{tabular}
	\end{spacing}
\end{table}

Forth, calculate the modulo values of different propositions according to the definition of QBPAs. All of the modulo values of separate propositions are given in Table 8.

\begin{table}[h]\footnotesize
\centering
\caption{Modulo values of different propositions}
\begin{spacing}{1.40}
	\begin{tabular}{c c c c c c}\hline
		$Modulo \ \ values$ &\multicolumn{5}{c}{$Values \ \ of \ \ propositions$}\\\hline
		&$\{x_{1}\}$&$\{x_{2}\}$&$\{x_{3}\}$&$\{x_{1}, x_{2}\}$&$\{\Theta\}$\\
		$\mathbb{Q}_{Modulo_{1}}$&$0.3770$&$0.4786$&$0.1443$&$1.8814e-05$&$2.7050e-06$\\\hline  
	\end{tabular}
\end{spacing}
\end{table}

Fifth, it can be easily obtained that proposition $x_{2}$ has the biggest modulo value after checking the data provided by Table 8. Therefore, the selected proposition is $x_{2}$.

Sixth, assume there is no threshold, then the objects, $\varpi$, which are corresponding to proposition $x_{2}$ are recognized as expected targets.
\clearpage
\subsection{Application of decision making}
TDQMF can not only play an important role in target recognition, but also have advantages in decision making which have been studied in many aspects \cite{Fei2019b,Xiao2019a,Yager2014a,Xiao2020a,Xiao2020c}. Nevertheless, how to make a reasonable decisions at the level of quantum is still a problem waiting to be solved. Compared with original quantum mass function, TDQMF has higher accuracy in indicating appropriate objects and is more efficient in making determinations. The following example illustrates the priority of TDQMF in decision making.
\subsubsection{Application 2}
Assume there is a securities company which makes decision about whether purchase a certain kind of stock or sell another category of stock. According to the situation of market, multiple situation awareness sensors give information in the form of quantum. The information is consist of two parts, the first part is the direct judgements made by sensors on whether purchasing a certain kind of stock or selling it. Except for those, an additional status is "doing nothing" which is also regarded as a kind of judgements. And the second part of information is the measure of the reliability of the first part given by another group of transducers.

Based on practical situations, some basic frames of this specific problem can be defined. The frame of discernment in the form of quantum of the first part is defined as $\Theta = \{P, S, D, PSD\}$. $P$ represents purchasing a kind of stock and $S$ denotes selling a kind of stock. Besides, $D$ means doing nothing and $PSD$ indicates that there is no idea about what to do. More than those, the frame of discernment in the form of quantum of the second part is defined as $\Theta = \{Y, N, YN\}$. Detailed data about the first dimension is given in table 9. Besides, detailed data of the second dimension are given in Table 10.

After the process of disposing, the data after modification is given in Table 11. Besides, the data after normalization is given in Table 12 and the data after combination is given in Table 13. In the last, the modulo values of different propositions are given in Table 14.
\begin{table}[h]\footnotesize
	\centering
	\caption{Quantum evidences given by multiple transducers}
	\begin{spacing}{1.40}
		\begin{tabular}{c c c c c}\hline
			$Quantum \ \ evidences$ &\multicolumn{4}{c}{$Values \ \ of \ \ propositions$}\\\hline
			&$\{P\}$&$\{S\}$&$\{D\}$&$\{PSD\}$\\
			$\mathbb{Q}_{1}^{1}$&$0.7810e^{0.8760j}$&$0.3605e^{0.5880j}$&$0.4382e^{0.4738j}$&$0.2605e^{0.6957j}$\\  
			$\mathbb{Q}_{1}^{2}$&$0.7366e^{0.8430j}$&$0.3640e^{0.6489j}$&$0.3848e^{0.4287j}$&$0.4204e^{0.3133j}$\\
			$\mathbb{Q}_{1}^{3}$&$0.8287e^{0.8451j}$&$0.3606e^{0.8050j}$&$0.3679e^{0.8238j}$&$0.2181e^{1.0946j}$\\
			$\mathbb{Q}_{1}^{4}$&$0.7789e^{0.8398j}$&$0.4177e^{0.8362j}$&$0.4318e^{0.7362j}$&$0.1793e^{0.5914j}$\\\hline
		\end{tabular}
	\end{spacing}
\end{table}
\begin{table}[h]\footnotesize
	\centering
	\caption{Another group of transducers' judgements on reliability of quantum evidences}
	\begin{spacing}{1.40}
		\begin{tabular}{c c c c c c}\hline
			$Quantum \ \ judgements$ &\multicolumn{4}{c}{$Values \ \ of \ \ propositions$}\\\hline
			&$\{Y\}$& $\{N\}$&$\{Y, N\}$&$\{\emptyset\}$\\
			$\mathbb{Q}_{2}^{1}$&$0.8275e^{0.7597j}$&$0.3905e^{0.6947j}$&$0.4032e^{0.5195j}$&$0$\\  
			$\mathbb{Q}_{2}^{2}$&$0.7920e^{0.8032j}$&$0.4177e^{0.8362j}$&$0.4450e^{0.7683j}$&$0$\\ 
			$\mathbb{Q}_{2}^{3}$&$0.7009e^{0.7349j}$&$0.4404e^{0.6889j}$&$0.5609e^{0.7511j}$&$0$\\
			$\mathbb{Q}_{2}^{4}$&$0.8287e^{0.7256j}$&$0.4263e^{0.6857j}$&$0.3623e^{0.1139j}$&$0$\\\hline   
		\end{tabular}
	\end{spacing}
\end{table}
\begin{table}[h]\footnotesize
	\centering
	\caption{Quantum evidences after modification}
	\begin{spacing}{1.40}
		\begin{tabular}{c c c c c}\hline
			$Combined \ \ TDQMF$ &\multicolumn{4}{c}{$Values \ \ of \ \ propositions$}\\\hline
			&$\{P\}$&$\{S\}$&$\{D\}$&$\{PSD\}$\\
			$\mathbb{Q}_{Modified_{1}}^{1}$&$1.0115e^{1.4983j}$&$0.8369e^{1.4011j}$&$0.8701e^{1.3761j}$&$0.5749e^{0.5725j}$\\
			$\mathbb{Q}_{Modified_{1}}^{2}$&$0.9885e^{1.5038j}$&$0.8509e^{1.4554j}$&$0.8545e^{1.4184j}$&$0.7075e^{0.5934j}$\\
			$\mathbb{Q}_{Modified_{1}}^{3}$&$0.9503e^{1.5674j}$&$0.8287e^{1.5496j}$&$0.8305e^{1.5519j}$&$0.6994e^{0.8211j}$\\
			$\mathbb{Q}_{Modified_{1}}^{4}$&$1.0548e^{1.5216j}$&$0.9099e^{1.5071j}$&$0.9165e^{1.4888j}$&$0.4716e^{0.2305j}$\\\hline  
		\end{tabular}
	\end{spacing}
\end{table}
\begin{table}[h]\footnotesize
	\centering
	\caption{Quantum evidences after normalization}
	\begin{spacing}{1.40}
		\begin{tabular}{c c c c c}\hline
			$Normalized \ \ quantum \ \ evidences$ &\multicolumn{4}{c}{$Values \ \ of \ \ propositions$}\\\hline
			&$\{P\}$&$\{S\}$&$\{D\}$&$\{PSD\}$\\
			$\mathbb{Q}_{Normalized_{1}}^{1}$&$0.6032e^{1.4983j}$&$0.4991e^{1.4012j}$&$0.5189e^{1.3762j}$&$0.3429e^{0.5725j}$\\  
			$\mathbb{Q}_{Normalized_{1}}^{2}$&$0.5773e^{1.5038j}$&$0.4969e^{1.4554j}$&$0.4990e^{1.4184j}$&$0.4132e^{0.5934j}$\\   
			$\mathbb{Q}_{Normalized_{1}}^{3}$&$0.5711e^{1.5675j}$&$0.4980e^{1.5496j}$&$0.4990e^{1.5520j}$&$0.4203e^{0.8211j}$\\ 
			$\mathbb{Q}_{Normalized_{1}}^{4}$&$0.6086e^{1.5216j}$&$0.5250e^{1.5072j}$&$0.5288e^{1.4888j}$&$0.2722e^{0.2305j}$\\\hline  
		\end{tabular}
	\end{spacing}
\end{table}
\begin{table}[h]\footnotesize
	\centering
	\caption{Quantum evidence after combination}
	\begin{spacing}{1.40}
		\begin{tabular}{c c c c c c}\hline
			$Combined \ \ quantum \ \ evidence$ &\multicolumn{4}{c}{$Values \ \ of \ \ propositions$}\\\hline
			&$\{P\}$&$\{S\}$&$\{D\}$&$\{PSD\}$\\
			$\mathbb{Q}_{Combined_{1}}$&$0.7050e^{0.8534j}$&$0.4887e^{0.6414j}$&$0.5134e^{0.6046j}$&$0.0211e^{1.5167j}$\\\hline  
		\end{tabular}
	\end{spacing}
\end{table}
\begin{table}[h]\footnotesize
	\centering
	\caption{Modulo values of different propositions}
	\begin{spacing}{1.40}
		\begin{tabular}{c c c c c c}\hline
			$Modulo \ \ values$ &\multicolumn{4}{c}{$Values \ \ of \ \ propositions$}\\\hline
	    	&$\{P\}$&$\{S\}$&$\{D\}$&$\{PSD\}$\\
			$\mathbb{Q}_{Modulo_{1}}$&$0.4970$&$0.2388$&$0.2636$&$0.0004$\\\hline  
		\end{tabular}
	\end{spacing}
\end{table}

It can be easily obtained that the quantum evidences given by sensors has conclusion that the securities company is expected to purchase this kind of stock. The operation which is supposed to be carried out is clear and decision-makers are able to distinguish the most reasonable way to manage current situations.
\clearpage
\subsection{Application of income estimate}
In the last two applications, the contents illustrate the advantages of TDQMF in target recognition and decision making. In this application, a more specific and practical case is offered to examine correctness and validity of TDQMF in ability of predictions based on actual conditions, namely income estimate. In fact, many researches on prediction about future trends which are even in the form of quantum have been made \cite{zhou2019robust, Dai2020, Xiao2020CEQD, Fan2019timeseries, Zhaojy2019timeseries}. Focused on the trend of changes of income, TDQMF is utilized to predict it based on provided effective and certain quantum evidences so as to give comparatively reasonable estimates.
\subsubsection{Application 3}
Assume there is a public company which needs a reasonable prediction about the expectation of future income. According to business circumstances and feedback of investment market, different sensors make their own judgements in the form of quantum about brief numbers of income the company is expected to receive. The circumstances of income of the company can be roughly divided into 3 ordered levels. Specific situations are "less than 1 billion", "between 1 billion and 10 billion" and "more than 10 billion". Propositions $\{\leq 1B, 1B-10B, \geq 10B, U\}$ represent the three status respectively. Therefore, the first dimension of frame of discernment can be defined as $\Theta = \{\leq 1B, 1B-10B, \geq 10B, U\}$. Among them, $U$ indicates that there is no predictions about the future income. And proposition $U$ can be also treated as an union set of proposition $\leq 1B$, $1B-10B$ and $\geq 10B$. Besides, the second dimension of the frame of discernment is defined as $\Theta = \{Y, N, YN\}$.

 Detailed information about the first dimension is given in table 15. Besides, detailed information of the second dimension is given in Table 16. More than that, the information after modification is given in Table 17. Except for those, the information after normalization is given in Table 18 and the combined information is given in Table 19. In the last, the modulo mass of different propositions is given in Table 20.
\begin{table}[h]\footnotesize
	\centering
	\caption{Quantum evidences given by multiple sensors}
	\begin{spacing}{1.40}
		\begin{tabular}{c c c c c}\hline
			$Quantum \ \ evidences$ &\multicolumn{4}{c}{$Values \ \ of \ \ propositions$}\\\hline
			&$\{\leq 1B\}$&$\{1B-10B\}$&$\{\geq 10B\}$&$\{U\}$\\
			$\mathbb{Q}_{1}^{1}$&$0.4472e^{1.1071j}$&$0.3905e^{0.8760j}$&$0.7810e^{0.8760j}$&$0.2397e^{0.5840}$\\  
			$\mathbb{Q}_{1}^{2}$&$0.4114e^{1.1180j}$&$0.4252e^{0.8519j}$&$0.7528e^{0.8794j}$&$0.2882e^{1.2164j}$\\
			$\mathbb{Q}_{1}^{3}$&$0.4429e^{1.0768j}$&$0.3764e^{0.6913j}$&$0.7798e^{0.8579j}$&$0.2321e^{0.8683j}$\\\hline
		\end{tabular}
	\end{spacing}
\end{table}
\begin{table}[h]\footnotesize
	\centering
	\caption{Another sensors' judgements on reliability of quantum evidences}
	\begin{spacing}{1.40}
		\begin{tabular}{c c c c c c}\hline
			$Quantum \ \ judgements$ &\multicolumn{4}{c}{$Values \ \ of \ \ propositions$}\\\hline
			&$\{Y\}$& $\{N\}$&$\{Y, N\}$&$\{\emptyset\}$\\
			$\mathbb{Q}_{2}^{1}$&$0.6640e^{0.9242j}$&$0.4924e^{1.1525j}$&$0.5626e^{0.6136j}$&$0$\\ 
			$\mathbb{Q}_{2}^{2}$&$0.7211e^{0.5880j}$&$0.5000e^{0.6435j}$&$0.4795e^{0.4586j}$&$0$\\
			$\mathbb{Q}_{2}^{3}$&$0.5336e^{0.2267j}$&$0.5580e^{0.6327j}$&$0.6389e^{0.2442j}$&$0$\\\hline  
		\end{tabular}
	\end{spacing}
\end{table}
\begin{table}[h]\footnotesize
	\centering
	\caption{Quantum evidences after modification}
	\begin{spacing}{1.40}
		\begin{tabular}{c c c c c}\hline
			$Combined \ \ TDQMF$ &\multicolumn{4}{c}{$Values \ \ of \ \ propositions$}\\\hline
			&$\{\leq 1B\}$&$\{1B-10B\}$&$\{\geq 10B\}$&$\{U\}$\\
			$\mathbb{Q}_{Modified_{1}}^{1}$&$0.9202e^{-1.1515j}$&$0.8982e^{-1.1348j}$&$0.9252e^{-1.1280j}$&$0.7223e^{0.6070j}$\\
			$\mathbb{Q}_{Modified_{1}}^{2}$&$0.9496e^{-1.5475j}$&$0.9493e^{1.5676j}$&$1.0215e^{1.5541j}$&$0.6329e^{0.6688j}$\\
			$\mathbb{Q}_{Modified_{1}}^{3}$&$0.9176e^{1.3873j}$&$0.8862e^{1.4104j}$&$0.8837e^{1.3070j}$&$0.7698e^{0.3619j}$\\\hline  
		\end{tabular}
	\end{spacing}
\end{table}
\begin{table}[h]\footnotesize
	\centering
	\caption{Quantum evidences after normalization}
	\begin{spacing}{1.40}
		\begin{tabular}{c c c c c}\hline
			$Normalized \ \ quantum \ \ evidences$ &\multicolumn{4}{c}{$Values \ \ of \ \ propositions$}\\\hline
			&$\{\leq 1B\}$&$\{1B-10B\}$&$\{\geq 10B\}$&$\{U\}$\\
			$\mathbb{Q}_{Normalized_{1}}^{1}$&$0.5285e^{-1.1515j}$&$0.5158e^{-1.1348j}$&$0.5313e^{-1.1280j}$&$0.4149e^{0.6070j}$\\  
			$\mathbb{Q}_{Normalized_{1}}^{2}$&$0.5269e^{-1.5475j}$&$0.5268e^{1.5676j}$&$0.5668e^{1.5541j}$&$0.3512e^{0.6688j}$\\   
			$\mathbb{Q}_{Normalized_{1}}^{3}$&$0.5296e^{1.3873j}$&$0.5115e^{1.4104j}$&$0.5101e^{1.3070j}$&$0.4443e^{0.3619j}$\\\hline  
		\end{tabular}
	\end{spacing}
\end{table}
\begin{table}[h]\footnotesize
	\centering
	\caption{Quantum evidence after combination}
	\begin{spacing}{1.40}
		\begin{tabular}{c c c c c c}\hline
			$Combined \ \ quantum \ \ evidence$ &\multicolumn{4}{c}{$Values \ \ of \ \ propositions$}\\\hline
			&$\{\leq 1B\}$&$\{1B-10B\}$&$\{\geq 10B\}$&$\{U\}$\\
			$\mathbb{Q}_{Combined_{1}}$&$0.5715e^{0.1205j}$&$0.5542e^{0.1467j}$&$0.5998e^{0.1749j}$&$0.0789e^{0.9842j}$\\\hline  
		\end{tabular}
	\end{spacing}
\end{table}
\begin{table}[h]\footnotesize
	\centering
	\caption{Modulo values of different propositions}
	\begin{spacing}{1.40}
		\begin{tabular}{c c c c c c}\hline
			$Modulo \ \ values$ &\multicolumn{4}{c}{$Values \ \ of \ \ propositions$}\\\hline
			&$\{\leq 1B\}$&$\{1B-10B\}$&$\{\geq 10B\}$&$\{U\}$\\
			$\mathbb{Q}_{Modulo_{1}}$&$0.3266$&$0.3072$&$0.3598$&$0.0062$\\\hline  
		\end{tabular}
	\end{spacing}
\end{table}

It can be easily concluded that the prediction system gives a estimate that the company is expected to receive a income which is higher than 10 billion. In this case, the ability to predict future trends of TDQMF is proved.
\clearpage
\subsection{Application of fault diagnosis}
In real life, how to correctly and accurately find errors existing in certain systems is a troublesome but crucial issue. As a result, fault diagnosis has been a heated topic in recent years, a lot of studies related with it have been made \cite{Zhang2020, Hang2014, Jiang2016, Basir2007, lin2018multisensor}. Most of them manage problems brought by fault diagnosis from the angle of Dempster-Shafer theory \cite{Dempster1967Upper, book}, certain specific faults can be diagnosed through customised methods. However, it is still vague and confusing when coming to seek for mistakes at the level of quantum. Therefore, usages in diagnosing different kinds of faults under complex situations are verified in this application.
\subsubsection{Application 4}
Assume there is a circuit board, one soldered dot decides whether the whole circuitry is interconnected and works normally. However, in the process of working, almost all of malfunctions is caused by the problems occurred in the soldered dot. Therefore, a group of sensors are arranged to report the judgements in the form of quantum on working status of the soldered dot. And the specific status can be described as "Normal", "faulted" and "sub-normal". Based on this practical situations, some basic circumstances of this specific problem can be depicted. The frame of discernment in the form of quantum of the first dimension is defined as $\Theta = \{Normal, Faulted, Sub-normal, W\}$. Among them, proposition $W$ indicates that there is no idea about the working status of the soldered dot. Besides, proposition $W$ can be also considered as a union set of the other three propositions, because $W$ represents an uncertainty in measuring actual situations. More than those, the frame of discernment in the form of quantum of the second dimension can be defined as $\Theta = \{Y, N, YN\}$. Detailed data about the first dimension is given in table 21. Besides, detailed data of the second dimension are given in Table 22.

After processing, the data which goes through modification is given in Table 23. Except for it, the data after normalization is given in Table 24 and the combined data is given in Table 25. In the last, the modulo values of different propositions are given in Table 26.
\begin{table}[h]\footnotesize
	\centering
	\caption{Quantum evidences given by multiple transducers}
	\begin{spacing}{1.40}
		\begin{tabular}{c c c c c}\hline
			$Quantum \ \ evidences$ &\multicolumn{4}{c}{$Values \ \ of \ \ propositions$}\\\hline
			&$\{Normal\}$&$\{Faulted\}$&$\{Sub-normal\}$&$\{W\}$\\
			$\mathbb{Q}_{1}^{1}$&$0.4964e^{1.0891j}$&$0.4219e^{1.0222j}$&$0.6560e^{0.9151j}$&$0.3808e^{0.6638j}$\\  
			$\mathbb{Q}_{1}^{2}$&$0.5069e^{1.1866j}$&$0.4579e^{1.0191j}$&$0.6312e^{0.8638j}$&$0.3671e^{0.3068j}$\\
			$\mathbb{Q}_{1}^{3}$&$0.4695e^{1.1071j}$&$0.4661e^{0.9530j}$&$0.6519e^{0.8505j}$&$0.3703e^{0.6713j}$\\
			$\mathbb{Q}_{1}^{4}$&$0.4632e^{1.0007j}$&$0.4455e^{0.8012j}$&$0.6801e^{0.8478j}$&$0.3525e^{0.4963j}$\\\hline
		\end{tabular}
	\end{spacing}
\end{table}
\begin{table}[h]\footnotesize
	\centering
	\caption{Another transducers' judgements on reliability of quantum evidences}
	\begin{spacing}{1.40}
		\begin{tabular}{c c c c c c}\hline
			$Quantum \ \ judgements$ &\multicolumn{4}{c}{$Values \ \ of \ \ propositions$}\\\hline
			&$\{Y\}$& $\{N\}$&$\{Y, N\}$&$\{\emptyset\}$\\
			$\mathbb{Q}_{2}^{1}$&$0.8000e^{0.3177j}$&$0.5793e^{0.3708j}$&$0.1555e^{0.6984j}$&$0$\\  
			$\mathbb{Q}_{2}^{2}$&$0.7564e^{0.4222j}$&$0.5232e^{0.4551j}$&$0.3923e^{0.5227j}$&$0$\\
			$\mathbb{Q}_{2}^{3}$&$0.7854e^{0.3781j}$&$0.5770e^{0.4868j}$&$0.2236e^{0.4636j}$&$0$\\
			$\mathbb{Q}_{2}^{4}$&$0.7430e^{0.5057j}$&$0.6352e^{0.4918j}$&$0.2104e^{0.6301j}$&$0$\\\hline   
		\end{tabular}
	\end{spacing}
\end{table}
\begin{table}[h]\footnotesize
	\centering
	\caption{Quantum evidences after modification}
	\begin{spacing}{1.40}
		\begin{tabular}{c c c c c}\hline
			$Combined \ \ TDQMF$ &\multicolumn{4}{c}{$Values \ \ of \ \ propositions$}\\\hline
			&$\{Normal\}$&$\{Faulted\}$&$\{Sub-normal\}$&$\{W\}$\\
			$\mathbb{Q}_{Modified_{1}}^{1}$&$1.1152e^{1.3251j}$&$1.0980e^{1.3202j}$&$1.1475e^{1.3013j}$&$0.4093e^{0.6768j}$\\
			$\mathbb{Q}_{Modified_{1}}^{2}$&$1.0663e^{1.3485j}$&$1.0054e^{1.3935j}$&$1.0438e^{1.3707j}$&$0.6336e^{0.4398j}$\\
			$\mathbb{Q}_{Modified_{1}}^{3}$&$1.1029e^{1.3867j}$&$1.0981e^{1.3729j}$&$1.1318e^{1.3490j}$&$0.4679e^{0.5727j}$\\
			$\mathbb{Q}_{Modified_{1}}^{4}$&$1.1367e^{1.3529j}$&$1.1362e^{1.3441j}$&$1.1613e^{1.3480j}$&$0.4443e^{0.5596j}$\\\hline  
		\end{tabular}
	\end{spacing}
\end{table}
\begin{table}[h]\footnotesize
	\centering
	\caption{Quantum evidences after normalization}
	\begin{spacing}{1.40}
		\begin{tabular}{c c c c c}\hline
			$Normalized \ \ quantum \ \ evidences$ &\multicolumn{4}{c}{$Values \ \ of \ \ propositions$}\\\hline
			&$\{Normal\}$&$\{Faulted\}$&$\{Sub-normal\}$&$\{W\}$\\
			$\mathbb{Q}_{Normalized_{1}}^{1}$&$0.5622e^{1.3251j}$&$0.5536e^{1.3202j}$&$0.5785e^{1.3014j}$&$0.2063e^{0.6768j}$\\  
			$\mathbb{Q}_{Normalized_{1}}^{2}$&$0.5589e^{1.3485j}$&$0.5270e^{1.3936j}$&$0.5472e^{1.3707j}$&$0.3321e^{0.4398j}$\\   
			$\mathbb{Q}_{Normalized_{1}}^{3}$&$0.5569e^{1.3867j}$&$0.5544e^{1.3729j}$&$0.5714e^{1.3491j}$&$0.2362e^{0.5727j}$\\ 
			$\mathbb{Q}_{Normalized_{1}}^{4}$&$0.5593e^{1.3529j}$&$0.5591e^{1.3441j}$&$0.5715e^{1.3480j}$&$0.2186e^{0.5596j}$\\\hline  
		\end{tabular}
	\end{spacing}
\end{table}
\begin{table}[h]\footnotesize
	\centering
	\caption{Quantum evidence after combination}
	\begin{spacing}{1.40}
		\begin{tabular}{c c c c c c}\hline
			$Combined \ \ quantum \ \ evidence$ &\multicolumn{4}{c}{$Values \ \ of \ \ propositions$}\\\hline
			&$\{Normal\}$&$\{Faulted\}$&$\{Sub-normal\}$&$\{W\}$\\
			$\mathbb{Q}_{Combined_{1}}$&$0.5787e^{0.9625j}$&$0.5453e^{0.9548j}$&$0.6062e^{0.9340j}$&$0.0064e^{-1.2502j}$\\\hline  
		\end{tabular}
	\end{spacing}
\end{table}
\begin{table}[h]\footnotesize
	\centering
	\caption{Modulo values of different propositions}
	\begin{spacing}{1.40}
		\begin{tabular}{c c c c c c}\hline
			$Modulo \ \ values$ &\multicolumn{4}{c}{$Values \ \ of \ \ propositions$}\\\hline
			&$\{Normal\}$&$\{Faulted\}$&$\{Sub-normal\}$&$\{W\}$\\
			$\mathbb{Q}_{Modulo_{1}}$&$0.3349$&$0.2974$&$0.3675$&$4.1071e-05$\\\hline  
		\end{tabular}
	\end{spacing}
\end{table}

It can be easily concluded that the status of the soldered dot is sub-normal. For workers, according to the outcomes of judgements, maintaining the dot is an important task to ensure the whole system is in a good condition. Therefore, TDQMF is able to find concrete faults and people could figure out corresponding solutions utilizing feedback from TDQMF. 
\clearpage
\section{Conclusion}
Quantum mass function proposes a novel angle to handle problems appeared in traditional evidence theory. However, the defects of original quantum mass function are also obvious. It may not reflect the actual conditions appropriately and give reasonable judgements. In order to solve these deficiencies, a two-dimensional quantum mass function is proposed in this paper. TDQMF not only takes actual situations into account, but also takes reliability of original judgements into consideration to give an authentic description of specific affairs. It can be regarded as a vigorous addition to quantum mass function which improves the confidential level and correctness of given results. Besides, the developed algorithm is also an efficient tool to extract truly useful key points in lots of applications when managing uncertain and complex quantum information. To sum up, TDQMF provides a feasible and applied solution to combine QBPAs and handle uncertainties existing in frames of discernment in the form quantum.
\section*{Acknowledgment}
 This research was funded by the Chongqing Overseas Scholars Innovation Program (No. cx2018077).
\bibliographystyle{elsarticle-num}
\bibliography{cite}
\end{document}